\documentclass[10pt,twocolumn,letterpaper]{article}

\usepackage[pagenumbers]{cvpr} 

\usepackage[dvipsnames]{xcolor}

\definecolor{cvprblue}{rgb}{0.21,0.49,0.74}
\usepackage[pagebackref,breaklinks,colorlinks,citecolor=cvprblue]{hyperref}
\usepackage{colortbl}

\def\paperID{*****} 
\def\confName{CVPR}
\def\confYear{2024}

\title{DEYOv3: DETR with YOLO for Real-time Object Detection}

\author{Haodong Ouyang\\
Southwest Minzu University\\
Chengdu, China\\
{\tt\small ouyanghaodong@stu.swun.edu.cn}
}

\begin{document}
\maketitle
\begin{abstract}
Recently, end-to-end object detectors have gained significant attention from the research community due to their outstanding performance. However, DETR typically relies on supervised pretraining of the backbone on ImageNet, which limits the practical application of DETR and the design of the backbone, affecting the model's potential generalization ability. In this paper, we propose a new training method called step-by-step training. Specifically, in the first stage, the one-to-many pre-trained YOLO detector is used to initialize the end-to-end detector. In the second stage, the backbone and encoder are consistent with the DETR-like model, but only the detector needs to be trained from scratch. Due to this training method, the object detector does not need the additional dataset (ImageNet) to train the backbone, which makes the design of the backbone more flexible and dramatically reduces the training cost of the detector, which is helpful for the practical application of the object detector. At the same time, compared with the DETR-like model, the step-by-step training method can achieve higher accuracy than the traditional training method of the DETR-like model. With the aid of this novel training method, we propose a brand-new end-to-end real-time object detection model called DEYOv3. DEYOv3-N achieves 41.1\% on COCO \texttt{val2017} and 270 FPS on T4 GPU, while DEYOv3-L achieves 51.3\% AP and 102 FPS. Without the use of additional training data, DEYOv3 surpasses all existing real-time object detectors in terms of both speed and accuracy. It is worth noting that for models of N, S, and M scales, the training on the COCO dataset can be completed using a single 24GB RTX3090 GPU. Code will be released at \url{https://github.com/ouyanghaodong/DEYOv3}.
\end{abstract}

\begin{figure}[t]
\vspace{3mm}
  \centering
  \includegraphics[width=0.9\linewidth]{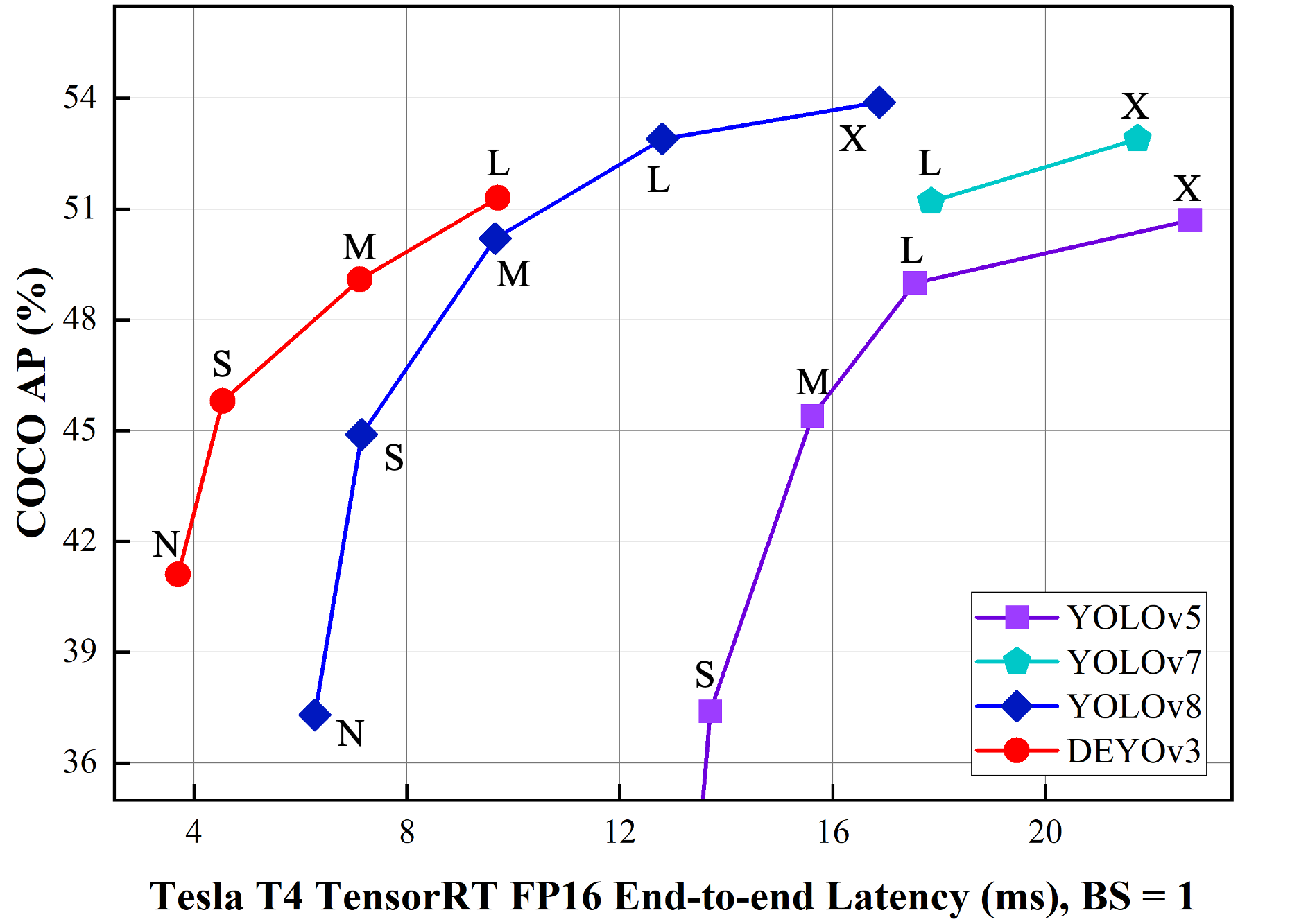}

   \caption{DEYOv3 surpasses other real-time object detectors in terms of speed and accuracy, establishing itself as a SOTA solution. Notably, all detectors were trained solely on the COCO dataset without using any additional datasets.}
   \label{fig:1}
\vspace{-5mm}
\end{figure}

\vspace{-4mm}
\section{Introduction}
\label{sec:intro}

\hspace{1pc}Object detection is a fundamental task in computer vision, aiming at accurately locating and recognizing multiple different classes of objects from images or videos. Object detection is the basis of many computer vision applications, including intelligent driving, video surveillance, face recognition, object tracking, etc. In recent years, deep learning methods, especially methods based on convolutional neural networks (CNN) \cite{64}, have made significant progress in object detection tasks and have become mainstream technical means. Real-time object detection is an essential topic in object detection, aiming at the task of quickly and accurately detecting and identifying objects in images or videos in real-time scenes. Compared with traditional object detection methods, real-time object detection requires faster processing speed and the ability to detect objects in real-time or near real-time. Existing real-time detectors generally adopt CNN-based architecture, which provides a good balance between accuracy and speed. Among them, one of the representatives of real-time detectors is YOLO \cite{4,5,6} (YOLO Only Look Once). After years of development, YOLO has developed into a series of fast models with good performance.

\begin{figure*}[t]
\begin{center}
\includegraphics[width=0.92\linewidth]{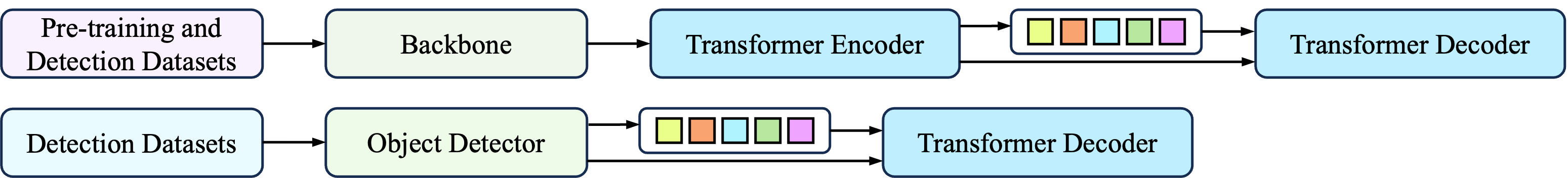}
   \caption{We showcase two distinct training strategies: DETR Strategy and Step-by-step Training. The DETR Strategy heavily relies on the ImageNet pre-trained backbone, while Step-by-step Training eliminates the need for additional datasets. Instead, it initializes DEYOv3's parameters by leveraging the pre-trained detector's backbone and multi-scale layers.}
\end{center}
\label{fig:2}
\vspace{-6.5mm}
\end{figure*}

Traditional detectors often require post-processing with non-maximum suppression (NMS). The effectiveness of NMS can be influenced by the chosen IoU threshold, which may lead to significant variations in the detection results. In crowded scenes, it can become a performance bottleneck for classical detectors and introduce inference latency for real-time detection. The Detection Transformer (DETR) \cite{1} proposes an innovative transformer-based \cite{10} object detector that leverages a transformer encoder-decoder framework that eliminates the manual components of NMS and instead utilizes the Hungarian loss to predict one-to-one sets of objects, enabling end-to-end optimization. Despite numerous works to improve DETR in recent years, the issue of high computational cost has remained unresolved, limiting the practical application and underutilization of its advantages. This means that although the object detection process is simplified, the high computational cost of the DETR model makes it challenging to achieve real-time object detection. \cite{58} re-evaluate DETR, reduce unnecessary computational redundancy in the DETR encoder, and propose the first end-to-end object detector, RT-DETR, fully exploiting the advantages of an end-to-end detection pipeline.

However, DETR \cite{1} typically relies on supervised pretraining of the backbone on ImageNet \cite{59}, as well as random initialization of the transformer encoder and decoder. If one intends to use a new backbone, a pre-trained backbone needs to be selected from ImageNet. Alternatively, after designing the backbone, it must be pre-trained on ImageNet before training DETR. This limits the design of the backbone and significantly increases training costs. Moreover, the performance and effectiveness of the model heavily depend on the dataset used for pretraining. If the current task's dataset differs significantly from the ImageNet, fine-tuning DETR may not fully adapt to specific tasks, resulting in degraded performance and limiting the robustness and generalization of DETR. To enhance the practicality of the DETR model, we propose a novel training method called step-by-step training. Specifically, in the first stage of training, we perform pre-training for one-to-many matching using YOLO \cite{4,5,6}. In the second stage of training, we utilize the backbone and neck of YOLO to initialize the backbone and encoder of the real-time end-to-end detector, while the decoder is randomly initialized for fine-tuning the one-to-one matching. Our training does not require additional datasets; only an object detection dataset is needed to complete the two stages of training. Additionally, due to the high-quality one-to-many matching pretraining performed by the multi-scale layers in the first stage, our method achieves higher accuracy without affecting inference time compared to the DETR training approach.

Furthermore, we propose a novel real-time object detection model, DEYOv3, based on a stepwise training approach. DEYOv3 eliminates the need for NMS, ensuring that the inference speed of the detector remains unaffected and stable. DEYOv3-N achieves 41.1\% AP on COCO \texttt{val2017} and runs at 270 FPS on NVIDIA Tesla T4 GPU, while DEYOv3-L achieves 51.3\% AP and 102 FPS. Without using additional training data, DEYOv3 outperforms all real-time detectors of similar scale in terms of both speed and accuracy, establishing itself as the new SOTA for real-time object detection.

The main contributions of this paper can be summarized as follows: (1) we propose a novel training method, step-by-step training, for the DETR model. Compared to traditional DETR training methods, it eliminates the need for additional datasets for pretraining and enables the model to achieve higher accuracy; (2) leveraging step-by-step training, we develop DEYOv3, the state-of-the-art real-time object detector; (3) we conduct a series of ablation experiments to thoroughly analyze DEYOv3 and discuss its potential as a feasible design approach for future large-scale object detection models.

\begin{figure*}[t]
\begin{center}
\vspace{-1.5mm}
\includegraphics[width=0.80\linewidth]{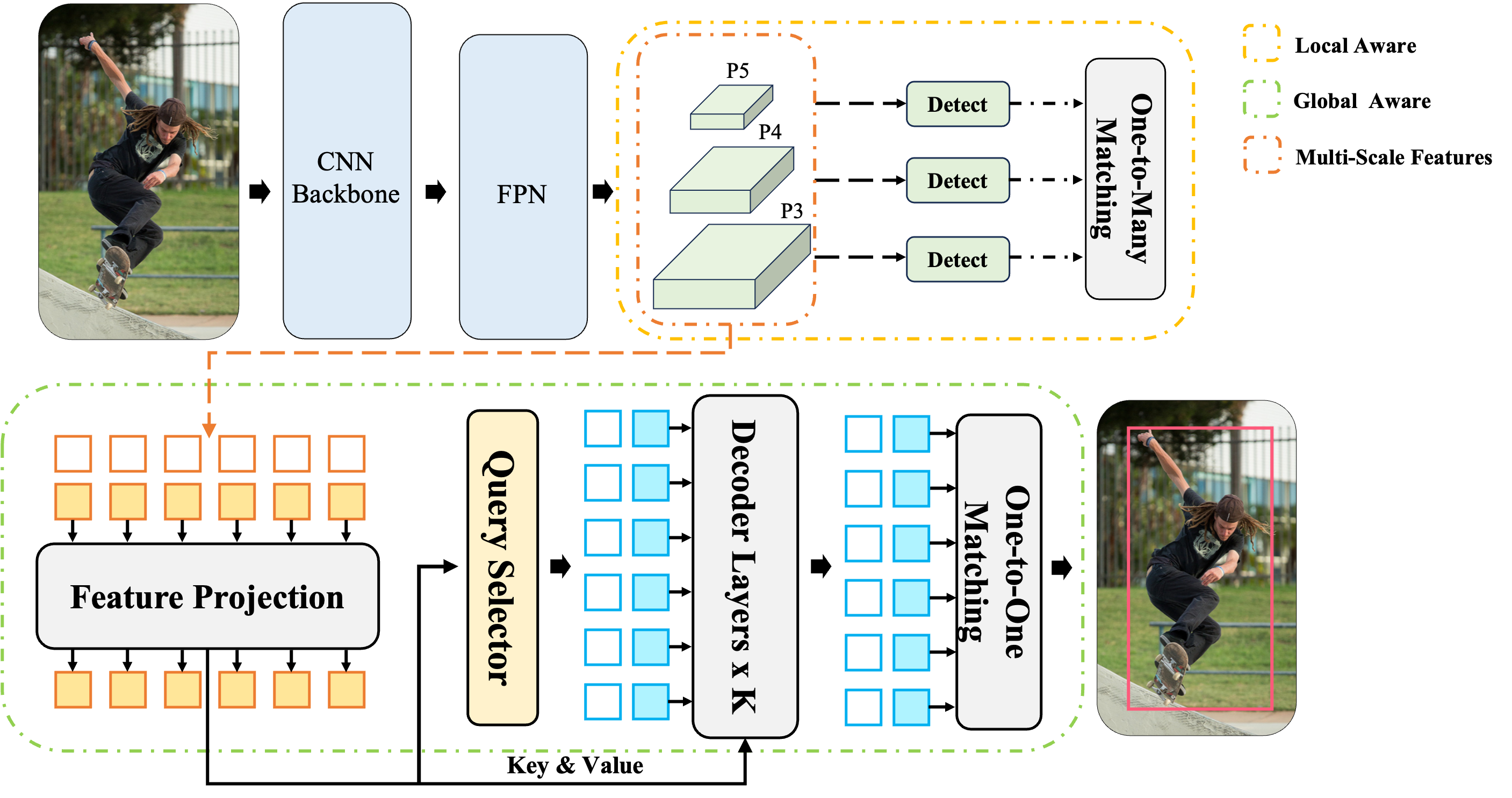}
   \caption{We eliminated the encoder usage and instead employed the multi-scale features \{P3, P4, P5\} provided by the neck. Following feature projection, these features were utilized as input for the encoder while simultaneously generating candidate bounding boxes and filtering them through the query selector. Subsequently, this information was passed into a decoder with an auxiliary prediction head, enabling iterative optimization for generating bounding boxes and scores.}
   \label{fig:3}
\end{center}
\vspace{-6.5mm}
\end{figure*}

\section{Step-by-step Training}

\hspace{1pc}Due to the Hungarian matching employed by DETR \cite{1} for directly predicting one-to-one sets of objects, as well as the quadratic relationship between the complexity of the DETR  decoder and sequence length, DETR achieves query numbers several orders of magnitude lower than classical detectors. We believe that this situation leads to relatively fewer supervisory signals during training for DETR, making it challenging to obtain good performance when training from scratch and thus highly reliant on ImageNet \cite{59} pretraining of the backbone. Furthermore, the multi-scale layers of the ImageNet pretraining have not been effectively pretrained specifically for the object detection task.

To address these issues, we divide the training process of DETR \cite{1} into two stages. Through extensive engineering tests, the computer vision community has thoroughly validated the generality and practicality of YOLO \cite{4,5,6}. These assessments include testing and evaluating YOLO in various real-world scenarios to assess its performance across different datasets, object categories, and environmental conditions. Based on these extensive engineering tests, we believe that even without using additional datasets, YOLO can still perform well in handling complex scenes, multi-object detection, and real-time applications while demonstrating excellent generalization and practicality. Undoubtedly, training YOLO from scratch is the optimal choice for its training strategy. Therefore, in the first stage of training, we initially train a powerful YOLO object detection model to provide high-quality pretraining for the backbone and multi-scale layers of the DEYOv3 model. Thanks to the one-to-many matching and the capability of YOLO to provide thousands of queries, a large amount of supervised signals enables the network to learn rich feature representations, thus providing better initial feature expression capabilities for the second stage of training. 

As the first stage provides high-quality pretraining for the backbone and neck of the DEYOv3 model, in the second stage of training, only the decoder needs to be trained from scratch, further accelerating the convergence speed of the model. Additionally, compared to the multi-scale layers pre-trained on ImageNet \cite{59}, the neck pre-trained in the first stage can provide higher-quality features. Compared to the traditional training method of DETR \cite{1}, our training strategy enables our model to exhibit better performance.
\section{DEYOv3}

\subsection{Model Overview}
\hspace{1pc}Based on the unreleased yolov8-rtdetr by Ultralytics \cite{62}, we construct the DEYOv3 model. As YOLO \cite{4,5,6} and DETR \cite{1} share the same backbone and neck, it is straightforward to present the DEYO \cite{52, 60} paradigm in a lightweight manner for real-time object detection. Figure~\ref{fig:3} illustrates the overall structure of DEYOv3. DEYOv3 utilizes YOLOv8 \cite{62} as a one-to-many branch of the model. YOLOv8 is an improved version of YOLOv5 \cite{61} introduced by Ultralytics, consisting of a backbone, a neck structure composed of Feature Pyramid Network (FPN) \cite{34} and Path Aggregation Network (PAN) \cite{35}, and a head that outputs predictions at three scales. The one-to-one branch of DEYOv3 employs a decoder similar to DINO, different from the previous DEYO \cite{52, 60} model, which does not use a transformer as an encoder like DETR. Instead, it encodes multi-scale features using a simple neck structure and feature projection method. The structure of feature projection is straightforward, consisting of a simple 1x1 convolution. This design makes DEYOv3 more lightweight during the encoding phase, improving the model's runtime efficiency. Additionally, DEYOv3's decoder adopts static query and dynamic initialization methods for anchor bounding boxes. Furthermore, DEYOv3 introduces an additional CDN (Contrastive Denoising Training) \cite{14} branch.

\begin{table*}[t]
\begin{center}
\resizebox{0.95\textwidth}{!}{
\begin{tabular}{lccccccccccc}
\hline
\bf{Model} &\bf{Backbone} &\bf{Epochs} &\bf{\#Params (M)} &\bf{GFLOPs} &\bf{FPS$_b$$_s$$_=$$_1$} & \bf{AP} & \bf{AP$_5$$_0$} & \bf{AP$_7$$_5$} & \bf{AP$_S$} & \bf{AP$_M$} & \bf{AP$_L$} \\
\hline
\rowcolor{gray!20}
\textit{Real-time Object Detectors} &\ &\ &\ &\ &\ &\ &\ &\ &\ &\ &\ \\
YOLOv5-N \cite{61} &-- &-- & 2 & 5 &76 &28.0 &46.2 &29.2 &14.1 &32.2 &36.7\\
YOLOv5-S \cite{61} &-- &-- & 7 & 17 &73 &37.4 &57.2 &40.2 &21.1 &42.3 &49.0\\
YOLOv5-M \cite{61} &-- &-- & 21 & 49 &64 &45.4 &64.4 &48.9 &27.8 &50.4 &58.1\\
YOLOv5-L \cite{61} &-- &-- & 47 & 109 &57 &49.0 &67.6 &53.1 &31.8 &54.4 &62.3\\
YOLOv5-X \cite{61} &-- &-- & 87 & 206 &44 &50.7 &68.9 &54.6 &33.8 &55.7 &65.0\\
YOLOv8-N \cite{62} &-- &-- & 3 &9 &159 &37.3 &52.5 &40.5 &18.6 &41.0 &53.5\\
YOLOv8-S \cite{62} &-- &-- & 11 &29 &139 &44.9 &61.8 &48.6 &25.7 &49.9 &61.0\\
YOLOv8-M \cite{62} &-- &-- & 26 &79 &103 &50.2 &67.2 &54.6 &32.0 &55.8 &66.4\\
YOLOv8-L \cite{62} &-- &-- & 44 &165 &78 &52.9 &69.8 &57.5 &35.3 &58.3 &69.8\\
YOLOv8-X \cite{62} &-- &-- & 68 &258 &59 &53.9 &71.0 &58.7 &35.7 &59.3 &70.7\\
\hline
\rowcolor{gray!20}
\textit{End-to-end Object Detectors} &\ &\ &\ &\ &\ &\ &\ &\ &\ &\ &\ \\
DETR \cite{1} &R50 &500 &41 &187 &-- &43.3 &63.1 &45.9 &22.5 &47.3 &61.1\\
Anchor-DETR \cite{46} &R50 &50 &39 &172 &-- &44.2 &64.7 &47.7 &23.7 &49.5 &62.3\\
Conditional-DETR \cite{22} &R50 &108 &44 &195 &-- &45.1 &65.4 &48.5 &25.3 &49.0 &62.2\\
Efficient-DETR \cite{47} &R50 &36 &35 &210 &-- &45.1 &63.1 &49.1 &28.3 &48.4 &59.0\\
SMCA-DETR \cite{38} &R50 &108 &40 &152 &-- &45.6 &65.5 &49.1 &25.9 &49.3 &62.6\\
Deformable-DETR \cite{7} &R50 &50 &40 &173 &-- &46.2 &65.2 &50.0 &28.8 &49.2 &61.7\\
DAB-Deformable-DETR \cite{8} &R50 &50 &48 &195 &-- &46.9 &66.0 &50.8 &30.1 &50.4 &62.5\\
DN-Deformable-DETR \cite{13} &R50 &50 &48 &195 &-- &49.5 &67.6 &53.8 &31.3 &52.6 &65.4\\
DINO \cite{14} &R50 &36 &47 &279 &5 &50.9 &69.0 &55.3 &34.6 &54.1 &64.6\\
\hline
\rowcolor{gray!20}
\textit{Real-time End-to-end Object Detectors} &\ &\ &\ &\ &\ &\ &\ &\ &\ &\ &\ \\
RT-DETR-R18 \cite{58} &R18 &72 &20 &60 &213 &46.5 &63.8 &-- &-- &-- &-- \\
RT-DETR-R34 \cite{58} &R34 &72 &31 &92 &154  &48.9 &66.8 &-- &-- &-- &-- \\
RT-DETR-R50 \cite{58} &R50 &72 &36 &100 &107  &53.1 &71.3 &57.7 &34.8 &58.0 &70.0 \\
RT-DETR-R101 \cite{58} &R101 &72 &42 &136 &70 &54.3 &72.7  &58.6 &36.0 &58.8 &72.1 \\
RT-DETR-L \cite{58} &HGNetv2 &72 &32 &110 &115 &53.0 &71.6 &57.3 &34.6 &57.3 &71.2 \\
RT-DETR-X \cite{58} &HGNetv2 &72 &67 &234 &73 &54.8 &73.1 &59.4 &35.7 &59.6 &72.9 \\
\rowcolor{red!10}
\textit{No Extra Training Data} &\ &\ &\ &\ &\ &\ &\ &\ &\ &\ &\ \\
DEYOv3-N &-- &72 &10 &17 &270 &41.1 &57.4 &44.2 &23.0 &44.2 &56.6\\
DEYOv3-S &-- &72 &16 &33 &220 &45.8 &63.1 &49.4 &27.3 &49.7 &61.4\\
DEYOv3-M &-- &72 &30 &76 &140 &49.1 &66.9 &53.2 &31.0 &53.3 &65.2\\
DEYOv3-L &-- &72 &46 &152 &102 &51.3 &69.1 &55.5 &35.4 &55.5 &66.1\\
\hline
\end{tabular}}
\end{center}
\vspace{-4mm}
\caption{Main results. Real-time detectors and our DEYOv3 utilize a consistent input size of 640, while end-to-end detectors employ an input size of (800, 1333). 
The end-to-end speed results are reported on a T4 GPU with TensorRT FP16, following the method proposed in RT-DETR. We do not test the speed of DETRs, as they are not real time detectors.}
\label{table:1}
\end{table*}

\subsection{One-to-many Branch}
\hspace{1pc}DEYOv3 adopts YOLOv8 \cite{62} as the one-to-many branch of the model to accommodate our step-by-step training method. YOLOv8 builds upon the success of previous YOLO \cite{4,5,6} versions and introduces new features and improvements to enhance performance and flexibility further. The three multi-scale layers of YOLOv8 provide the one-to-one branch with up to 8400 queries, which can be used to generate proposal bounding boxes and serve as the key and value for the decoder. Unlike DETR \cite{1}, YOLO benefits from the one-to-many training approach, which allows these queries to be more thoroughly supervised during the first-stage training. As a result, a powerful neck is trained to provide multi-scale information to the decoder, enabling the model to achieve better performance.

\subsection{Efficient Encoder}
\hspace{1pc}The encoder of DEYOv3 differs from DETR in that it does not employ a transformer as the encoder. Instead, DEYOv3 utilizes YOLO's pre-trained neck in the first stage to encode multi-scale features, which are then fed into feature projection to align them to the hidden dimension. In a broad sense, the entire neck and feature projection can be considered as the encoder of DEYOv3. This implementation aligns well with our step-by-step training method. Due to the rich features obtained by the neck during efficient pre-training in the first stage, these features can provide efficient initialization for the encoder in the second stage. As a result, the encoder can offer high-quality key, value, and proposal bounding box information to the decoder. Compared to DETR's randomly initialized multi-scale layers and encoder, DEYOv3's encoder achieves exceptional speed while simultaneously maintaining performance. We can describe this process as follows:
\begin{align}
\nonumber
& \textcolor{black}{S_1 = Proj (P_3, P_4, P_5)}\\
& \textcolor{black}{S_2 = Concat(S_1)}\\
\nonumber
& \textcolor{black}{Q = K = V = S_2}
\end{align}

\subsection{One-to-one Branch}
\hspace{1pc}The decoder of DEYOv3 adopts a similar architecture to DETR \cite{1}, utilizing self-attention in transformers \cite{10} to capture relationships between different queries, thereby establishing score disparities to suppress redundant bounding boxes. In each level of the decoder, the queries are progressively refined, resulting in a one-to-one set of objects. This design greatly simplifies the object detection process in DEYOv3 and eliminates the reliance on non-maximum suppression (NMS). Additionally, thanks to the global awareness provided by the transformer decoder, similar to DETR, DEYOv3 exhibits improved classification capability. Furthermore, due to the high-quality initialization provided by the first-stage training, even with supervision on only a few hundred queries in the one-to-one branch, the model can converge rapidly and achieve better performance.

\begin{figure}[t]
  \centering
  \includegraphics[width=0.9\linewidth]{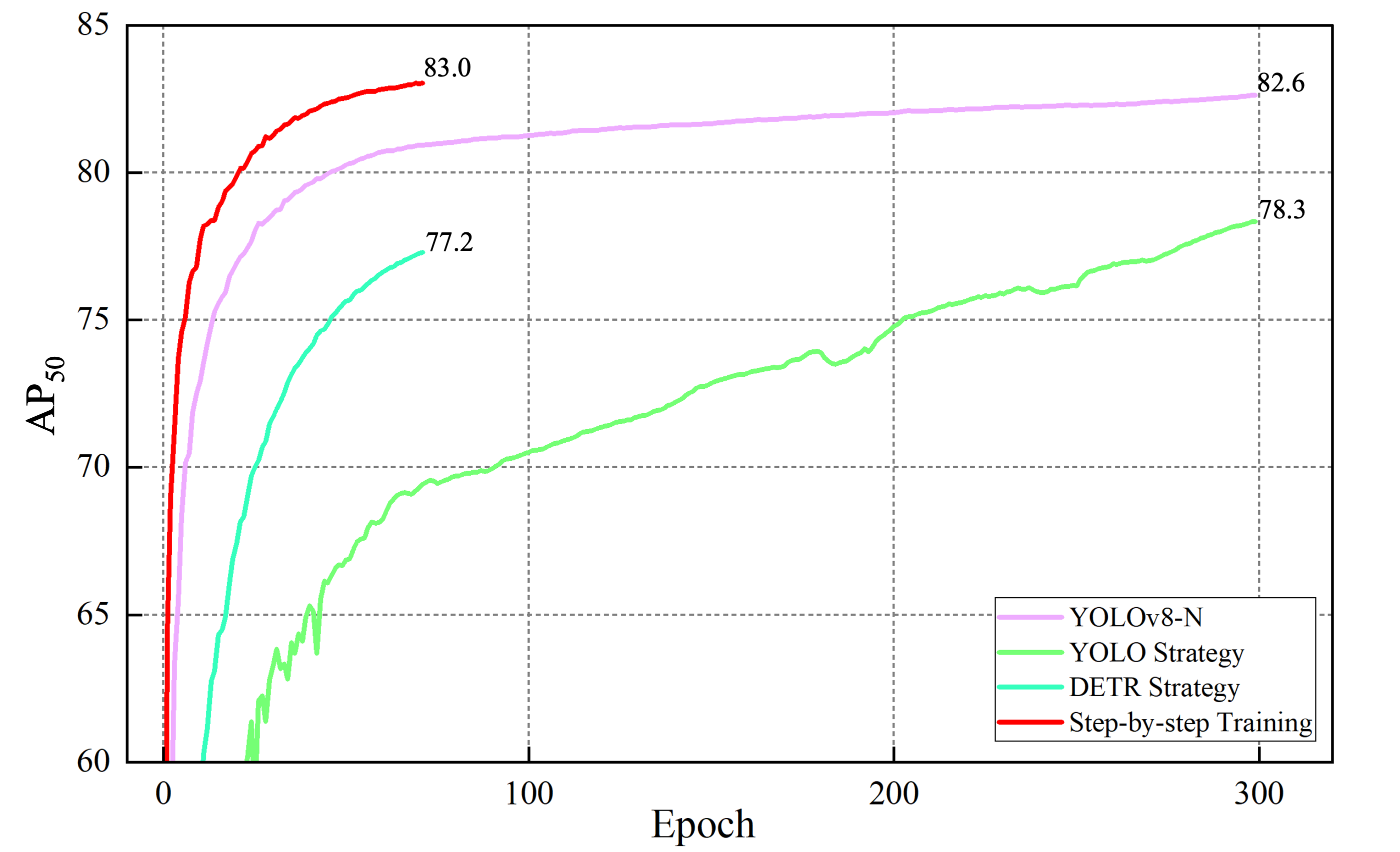}
   \vspace{-1.5mm}
   \caption{Convergence curves of various methods trained on CrowdHuman. Please note that we calculate the AP$_5$$_0$ using the tools of YOLOv8 in this experiment.}
   \label{fig:4}
\vspace{-5.5mm}
\end{figure}

\section{Experiments}
\subsection{Setups}

\noindent {\bf COCO} To evaluate the performance of our method in object detection tasks, we conducted experiments on the widely used Microsoft COCO \cite{12}. We trained the DEYOv3 using the \texttt{train2017} and evaluated the performance using the \texttt{val2017}. In our experiments, we did not utilize any additional training data.

\noindent {\bf CrowdHuman} To evaluate the end-to-end effectiveness of DEYOv3 in dense detection compared to classical detectors, we conducted experiments on CrowdHuman \cite{63}. The CrowdHuman consists of 15,000 images for training, 4,370 images for validation, and 5,000 images for testing. We utilized the provided full-body annotations in the dataset and evaluated on the validation set. In terms of optimizer-related parameters, we adopted the same settings as the COCO.

\noindent {\bf Implementation Details} In the first stage of training DEYOv3, we used a 6-layer transformer decoder with hidden feature dimensions of 256. We trained the detector from scratch following \cite{62} strategy and hyperparameters. In the second stage of training, we further trained our detector using the AdamW \cite{33} optimizer. The learning rate was set to 0.0001, the learning rate for the backbone was set to 0.00001, and the weight decay was set to 0.0001. The data augmentation strategy in the second stage was similar to the first stage, including random color distortion, translation, flipping, resizing, and other operations. However, unlike the first stage of data augmentation, in the second stage of training, we turned off mosaic data augmentation and stopped using gray filling to pad image borders.

\begin{table}[h]
\vspace{1.5mm}
\begin{center}
\resizebox{0.43\textwidth}{!}{
\begin{tabular}{lccccc}
\hline
Model &Epochs &AP$_5$$_0$ &mMR &Recall \\
\hline
YOLOv8-N &300 &83.0 &49.9 &84.6 \\
\rowcolor{gray!10}
DEYOv3-N &72 &87.1 &49.3 &95.0 \\
\hline
\end{tabular}}
\end{center}
\vspace{-3mm}
\caption{Perfermance on CrowdHuman (full body).}
\vspace{-2mm}
\label{table:2}
\end{table}

\subsection{Main Results}
\hspace{1pc}We compared the scaled DEYOv3 with YOLOv5 \cite{61}, YOLOv8 \cite{62}, and RT-DETR \cite{58} in Table~\ref{table:1}. Compared to YOLOv8, DEYOv3 demonstrates a significant improvement in accuracy by 3.8\% / 0.9\% AP at scales N and S while achieving a 70\% / 59\% increase in FPS. At scales M and L, DEYOv3 continues to exhibit a better trade-off between accuracy and speed. Additionally, compared to non-real-time end-to-end detectors, DEYOv3 showcases remarkable speed advantages. Meanwhile, DEYOv3-N has demonstrated a remarkable increase of 4.1 AP compared to YOLOv8-N in dense detection scenarios of CrowdHuman \cite{63}.

\begin{figure*}[h]
\begin{center}
\includegraphics[width=0.95\linewidth]{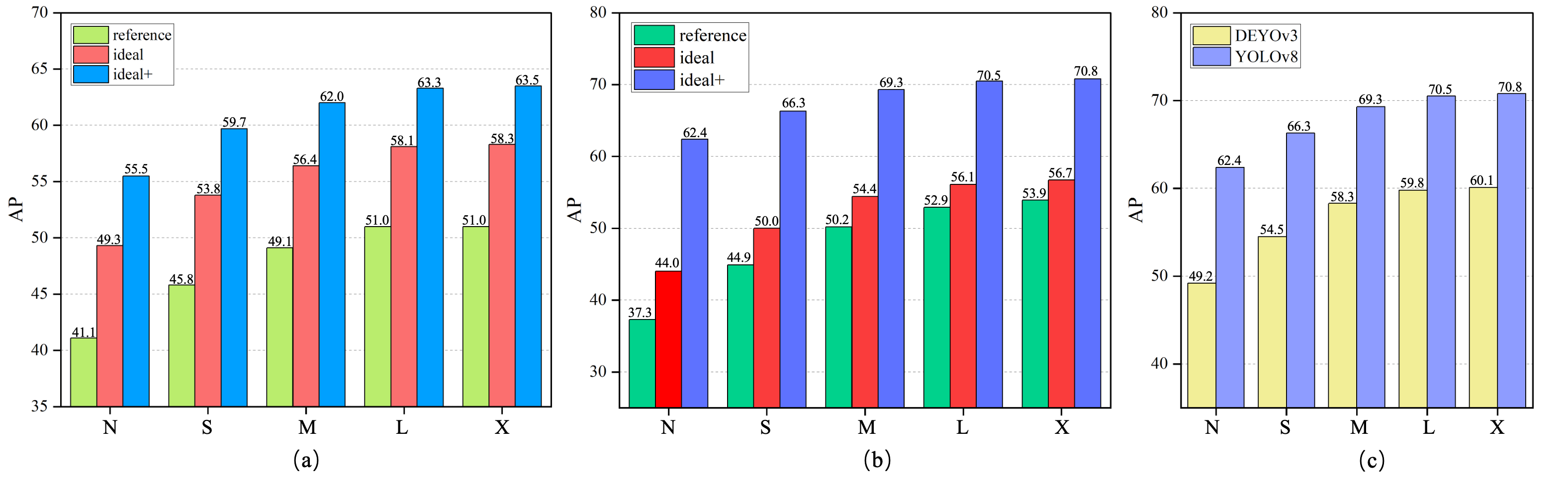}
\vspace{-2.5mm}
   \caption{The exploration of the ideal performance of YOLOv8 and DEYOv3. Here, reference represents the model's performance under standard inference, while ideal represents where we perform one-to-one matching between labels and inference results to distinguish true positives (TP) from false positives (FP), thereby achieving the purpose of measuring the ideal performance of the model. Ideal+ builds upon ideal by eliminating classification errors and further examines the model's ability to localize objects. Experiments were conducted on DEYOv3 and YOLOv8 in Figures (a) and (b), respectively. Figure (c) compares the capabilities of the one-to-many branch and one-to-one branch in generating candidate boxes. In (a), the bounding box from the last layer of the decoder in DEYOv3 was used, while in (c), the bounding boxes of DEYOv3 were derived from the bounding boxes after being filtered through the query selector.}
   \label{fig:5}
\end{center}

\vspace{-5.5mm}
\end{figure*}

\subsection{Ablation Study}

\hspace{1pc}In Fig.~\ref{fig:4}, we present the convergence curves of YOLOv8-N trained from scratch on CrowdHuman \cite{63}, as well as DEYOv3-N trained with three different training strategies: YOLO Strategy, DETR Strategy, and Step-by-step Training. Even with training from scratch, YOLOv8-N \cite{62} achieves the good performance of 82.6 AP without relying on additional datasets, thanks to the rich supervision signals provided by one-to-many training. However, DEYOv3 trained with the same training strategies and limited supervision signals due to one-to-one matching only achieves a performance of 78.3 AP. Furthermore, initializing DEYOv3's backbone with YOLOv8-N-CLS \cite{62} pre-trained on ImageNet and using the training strategy of DETR only yields a performance of 72.1 AP. 

\begin{table}[h]
\begin{center}
\resizebox{0.43\textwidth}{!}{
\begin{tabular}{lccccc}
\hline
Model &Epochs & Backbone & Neck &AP$_5$$_0$ \\
\hline
DEYOv3-N &72 &\checkmark & &68.3\\
DEYOv3-N &72 &\checkmark &\checkmark &\bf{87.1}\\
\hline
\end{tabular}}
\end{center}
\vspace{-4mm}
\caption{Results of the ablation study on step-by-step training.}
\label{table:3}
\vspace{-2mm}
\end{table}
 
In Table~\ref{table:3}, we analyze the necessity of the high-quality multi-scale features provided by the neck pre-trained in the first stage. When using the backbone with step-by-step training without initializing the neck, the model's performance decreases significantly by 18.8 AP, reaching a minimum of 68.3 AP. This clearly demonstrates that the key to achieving good performance in DEYOv3 lies not in a more powerful backbone compared to ImageNet \cite{59} pretraining but instead in the utilization of a Neck pre-trained in the first stage, which provides high-quality multi-scale features for the decoder.

\subsection{Analysis}

\hspace{1pc}In Table~\ref{table:4}, we observed that the model scaling strategy of YOLOv8 \cite{62} is not applicable to DEYOv3, as the output dimensions of the Neck do not match the hidden dimensions of the decoder. The use of an X-scale model as the one-to-many branch in DEYOv3 not only slows down the inference speed but also fails to improve the AP. Although DEYOv3 falls short of achieving optimal performance at larger scales, this actually highlights the tremendous potential of the DEYOv3 model. We believe that a specifically designed one-to-many branch or a more powerful feature projection module can elevate the performance of DEYOv3 to new heights.

\begin{table}[h]
\begin{center}
\resizebox{0.45\textwidth}{!}{
\begin{tabular}{lccccc}
\hline
Model Size &Neck &Hidden Dimension &FPS$_b$$_s$$_=$$_1$ &AP \\
\hline
L  &(256, 512, 512)  &256 &102 &51.3\\
L  &(256, 512, 512)  &512 &85  &51.6\\
X  &(256, 512, 1024) &256 &75 &51.3\\
\hline
\end{tabular}}
\end{center}
\vspace{-4mm}
\caption{The ablation study on model scaling, we demonstrate the impact of different output dimensions of multi-scale layers and hidden layer dimensions on the final performance.}
\vspace{-3mm}
\label{table:4}
\end{table}

According to the results shown in Fig.~\ref{fig:5}, we observed that DEYOv3 outperforms YOLOv8 \cite{62} in terms of performance. This can be attributed to the stronger classification capability of DEYOv3's decoder. Although YOLOv8 \cite{62} can provide higher-quality candidate boxes, it is challenging to translate them into effective performance. In other words, DEYOv3's decoder exhibits a higher ability to improve classification accuracy, thereby achieving better performance in object detection tasks.

\begin{table}[h]
\vspace{1.5mm}
\begin{center}
\resizebox{0.43\textwidth}{!}{
\begin{tabular}{lccccc}
\hline
Model &Epochs &AP$_5$$_0$ &mMR &Recall \\
\hline
DEYOv3-N &72 &87.1 &49.3 &95.0 \\
\rowcolor{gray!10}
DEYOv3-N+ &300 &86.7 &49.3 &93.4 \\
\hline
\end{tabular}}
\end{center}
\vspace{-4mm}
\caption{The comparison of results between DEYOv3 and its improved version DEYOv3+ on CrowdHuman.}
\label{table:5}
\end{table}

\begin{figure*}[h]
\begin{center}
\includegraphics[width=0.9\linewidth]{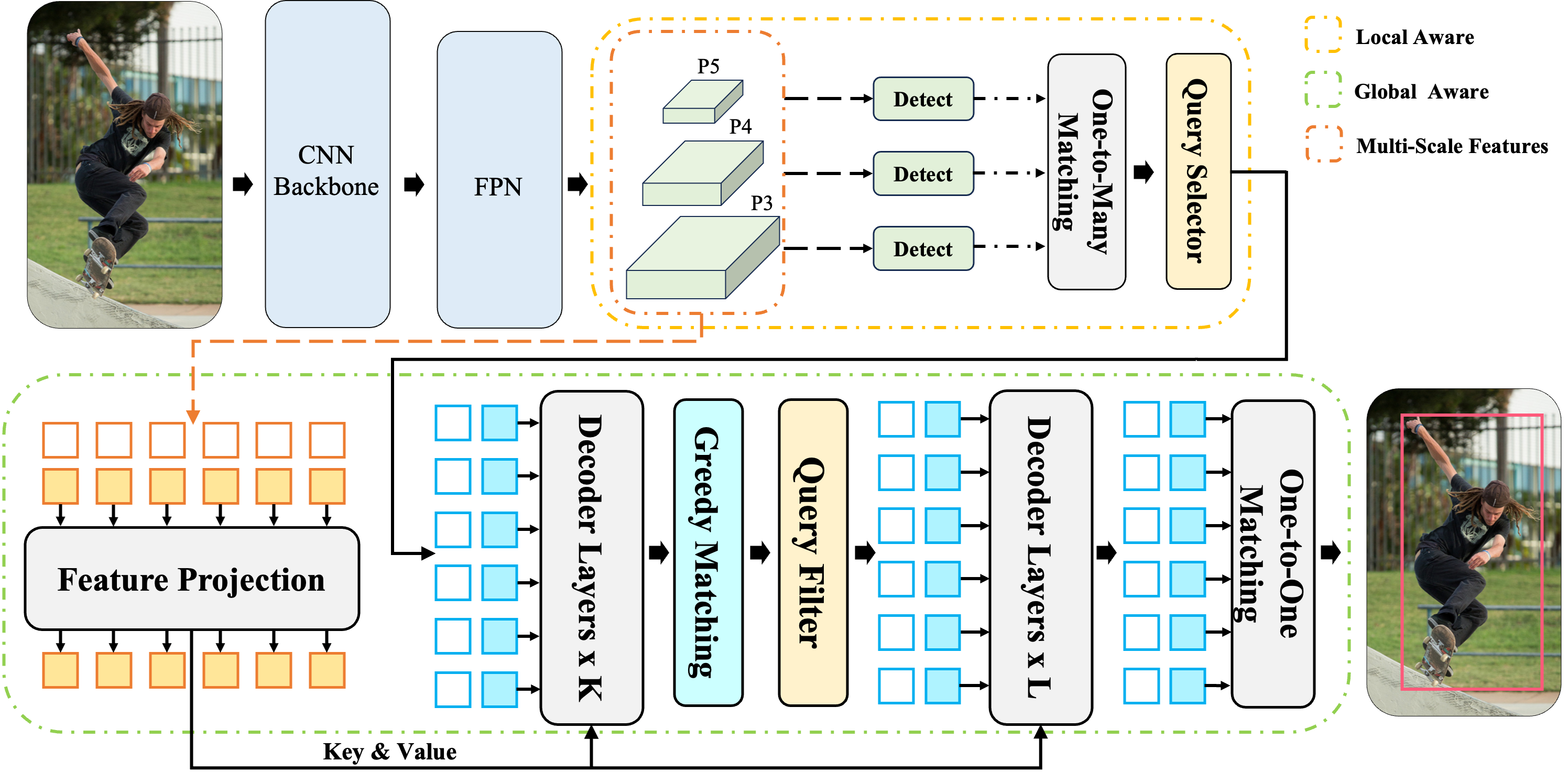}
   \caption{The architecture of the improved DEYOv3 model follows the design of DEYOv2, which allows the decoder to obtain higher-quality proposal bounding boxes.}
   \label{fig:6}
\end{center}
\vspace{-6mm}
\end{figure*}

\begin{table}[h]
\begin{center}
\resizebox{0.45\textwidth}{!}{
\begin{tabular}{lcccccccc}
\hline
Row & Model &epochs & o2m & o2o & pad &AP$_{50(o2m)}$ &AP$_{50(o2o)}$ \\
\hline
1 & YOLOv8-N &300 & \checkmark & &\checkmark & 83.0 &-- \\
2 & YOLOv8-N &300 & \checkmark & & & 83.0 &-- \\
3 & DEYOv3-N &72 & &\checkmark &\checkmark &-- &86.7 \\
4 & DEYOv3-N &72 & & \checkmark & &--  &87.1 \\
5 & DEYOv3-N &72 &\checkmark &\checkmark & &84.1 &86.7 \\
6 & \hspace{0.5em}DEYOv3-N+ &300 &\checkmark & \checkmark & & 84.2 &86.7 \\
\hline
\end{tabular}}
\end{center}
\vspace{-2.5mm}
\caption{More ablation study on DEYOv3, where pad refers to filling the image borders with gray color.}
\vspace{-2.5mm}
\label{table:6}
\end{table}

\subsection{Revisiting the DEYO}
\hspace{1pc}To enhance the quality of candidate boxes generated by DEYOv3's decoder, we further integrated DETR with YOLO. Specifically, the improved DEYOv3 incorporates a one-to-many branch to generate candidate boxes. Additionally, we introduced DEYOv2's rank feature \cite{60}, greedy matching, and query filter. Apart from architectural improvements, we conducted joint training in the second stage to ensure that the one-to-many branch provides high-quality candidate boxes. We transformed the original prediction for one-to-one object sets into a multi-objective optimization problem. Specifically, while using one-to-one matching to supervise the one-to-one branch, we employed a loss function based on one-to-many matching to supervise the one-to-one branch. This process can be described as follows:
\begin{align}
& \textcolor{black}{L_{total} = L_{o2m} + L_{o2o}}
\end{align}

However, despite our efforts to improve DEYOv3, its performance did not reach the desired level compared to DEYOv2 \cite{60}. This may be attributed to the insufficient feature representation provided without the use of an encoder in DEYOv3, which fails to suppress redundant bounding boxes effectively. As shown in Fig.~\ref{fig:6}, the model's improvement on candidate boxes is limited at larger scales, while the enhancement in classification capability surpasses the improvement in candidate box quality. We believe that with a well-designed scale, a balance can be achieved between the quality of candidate boxes and the strength of classification features, thereby overcoming the bottleneck caused by candidate boxes and further enhancing the performance of DEYOv3.

We conducted additional ablation experiments in Table~\ref{table:6}. The results in rows 1, 2, and 3 indicate that training the o2m branch without using gray padding to fill the image borders ultimately improves the performance of DEYOv3, suggesting significant room for improvement in DEYOv3 on COCO. The results in rows 4, 5, and 6 demonstrate that joint training can effectively accommodate both the one-to-one branch and the one-to-many branch.

\begin{figure}[h]
  \centering
  \includegraphics[width=0.7\linewidth]{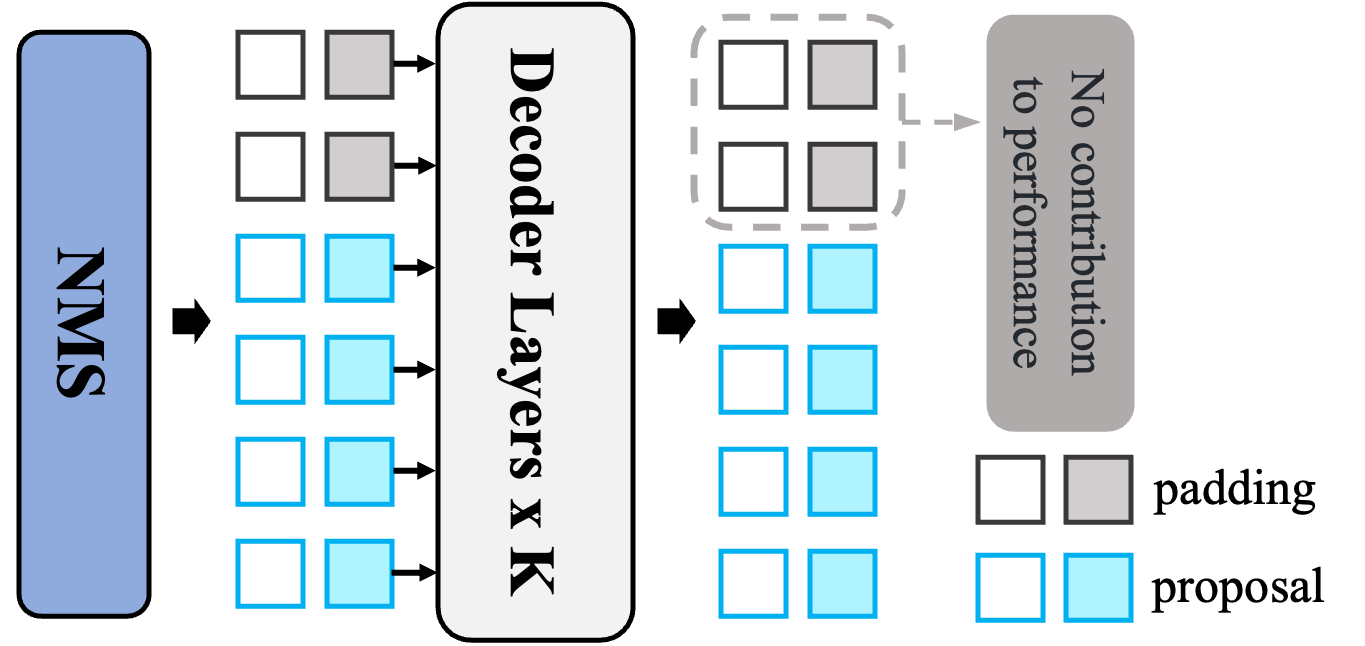}
   \vspace{-1.5mm}
   \caption{Padding queries do not contribute anything to the final performance.}
   \label{fig:7}
\vspace{-5.5mm}
\end{figure}

Why don't we use NMS to suppress redundant bounding boxes, which can significantly reduce the requirements for the model's classification capability? However, we found that this approach still limits the detector's performance due to the bottleneck caused by NMS. As shown in Fig.~\ref{fig:7}, We found that the padded queries contribute nothing to the final results, indicating that the candidate boxes filtered by NMS are effective queries. In the decoding stage, there is no way to compensate for the performance loss caused by NMS incorrectly removing boxes. Furthermore, the latency of NMS inference is unstable, which significantly affects the speed of the detector. Query filter can avoid these issues and potentially achieve ideal+ theoretical performance under ideal conditions.

\section{Related Work}
\label{sec:formatting}

\begin{figure*}[t]
\begin{center}
\includegraphics[width=0.9\linewidth]{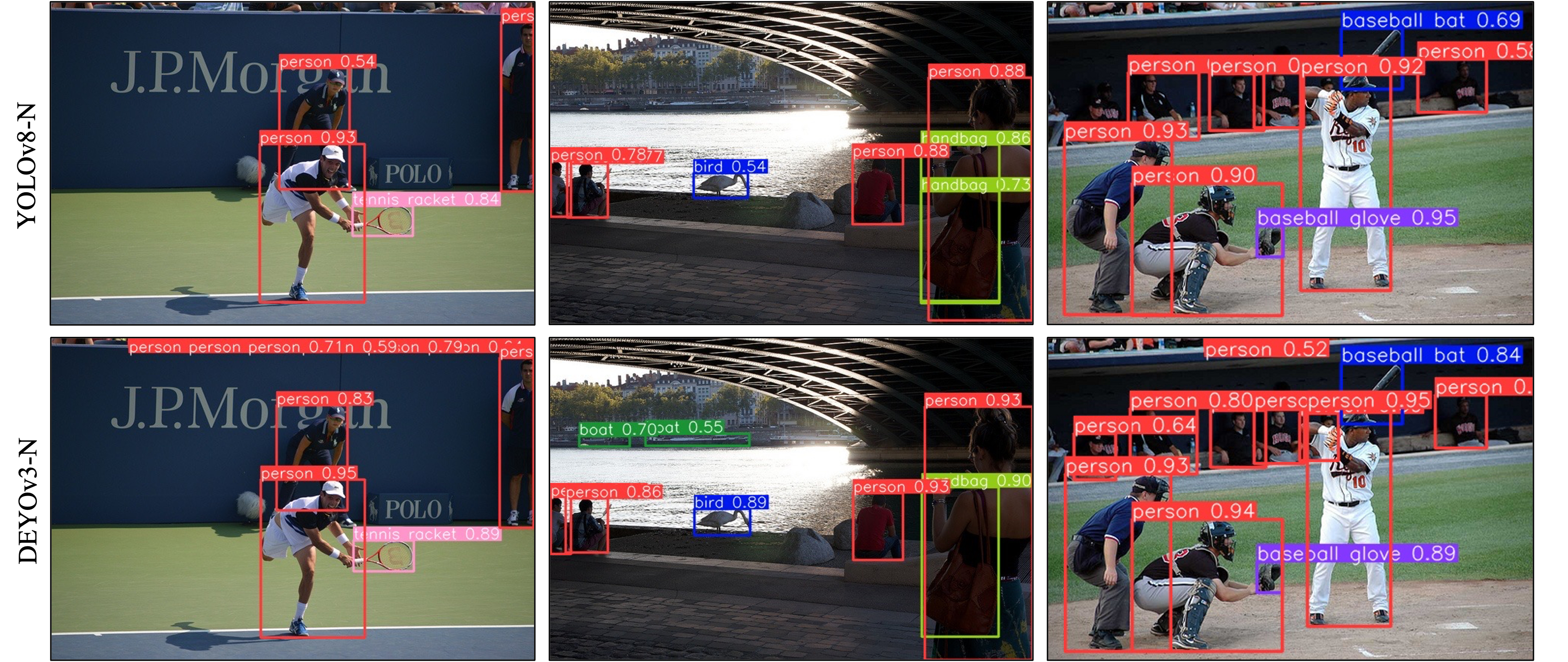}
   \caption{As shown in the figure, from the results, it intuitively shows that DEYOv3-N achieves better performance with lower computational cost.
}
\end{center}
\label{fig:8}
\vspace{-6mm}
\end{figure*}

\subsection{You Only Look Once}
\hspace{1pc}Over the years, the YOLO \cite{4,5,6} series has been one of the best single-stage real-time object detector categories. YOLO transforms the object detection task into a regression problem, predicting the positions and categories of multiple objects in a single forward pass, achieving high-speed object detection. After years of development, YOLO has developed into a series of fast models with good performance. Anchor-based YOLO methods include YOLOv4 \cite{25}, YOLOv5 \cite{61}, and YOLOv7 \cite{26}, while anchor-free methods are YOLOX \cite{27}, YOLOv6 \cite{29}, and YOLOv8 \cite{62}. Considering the performance of these detectors, anchor-free methods perform as well as anchor-based methods, and anchor boxes are no longer the main factor limiting the development of YOLO. However, all YOLO variants generate many redundant bounding boxes, which NMS must filter out during the prediction stage, which has a significant impact on the accuracy and speed of the detector and conflicts with the design theory of real-time object detectors.

\subsection{End-to-end Object Detectors}
\hspace{1pc}Carion et al. proposed an end-to-end object detector based on transformers, named DETR (DEtection TRansformer) \cite{1}, which has attracted significant attention from researchers due to its end-to-end nature in object detection. Specifically, DETR eliminates the anchor and NMS components in traditional detection pipelines and adopts a bipartite graph matching label assignment method to directly predict one-to-one sets of objects. This strategy greatly simplifies the object detection process and alleviates the performance bottleneck caused by NMS. However, DETR suffers from slow convergence speed and query ambiguity issues. To address these problems, several variants of DETR have been proposed, such as Deformable-DET \cite{7}, Conditional-DETR \cite{22}, Anchor-DETR \cite{46}, DAB-DETR \cite{8}, DN-DETR \cite{13}, and DINO \cite{14}. Deformable-DETR enhances the efficiency of attention mechanisms and accelerates training convergence by utilizing multi-scale features. Conditional-DETR and Anchor-DETR reduce the optimization difficulty of queries. DAB-DETR introduces 4D reference points and optimizes predicted boxes layer by layer. DN-DETR speeds up training convergence by introducing query denoising. DINO improves upon previous work and achieves state-of-the-art results. However, the aforementioned improvements do not address the issue of high computational cost in DETR. RT-DETR \cite{58} designs an efficient hybrid encoder to replace the original transformer encoder, reducing unnecessary computational redundancy in the DETR encoder and proposing the first end-to-end object detector.

\subsection{DETR with YOLO}
\hspace{1pc}DEYO (DETR with YOLO) addresses the issues of DETR from a novel perspective. By combining the strengths of both classical detectors and the DETR-like model, DEYO \cite{52} achieved state-of-the-art performance at the time. However, DEYO heavily relies on NMS to filter out redundant boxes from the classical detector to avoid impacting the optimization of one-to-one matching. DEYOv2 \cite{60} overcomes this dependency on NMS by adopting greedy matching and ranking features. It is the first complete end-to-end model that combines the advantages of classical detectors and query-based object detection models.

\section{Conclusion}
\hspace{1pc}We propose a new training method called step-by-step training, and leveraging this method, we introduce a novel real-time end-to-end detector named DEYOv3. DEYOv3 surpasses all existing real-time detectors without the need for additional training data. 

We hope this work will find wide application in practical scenarios and inspire researchers. However, this novel detector design also brings new challenges, particularly in large-scale model design and overcoming localization deficiencies. We believe that the DEYOv3 model has tremendous potential, similar to DETR \cite{1}, and we look forward to future work successfully addressing these challenges faced by DEYOv3.

{
    \small
    \bibliographystyle{ieeenat_fullname}
    \bibliography{main}
}


\end{document}